\documentclass[10pt,twocolumn,letterpaper]{article}

\usepackage{iccv}
\usepackage{times}
\usepackage{epsfig}
\usepackage{graphicx}
\usepackage{amsmath}
\usepackage{amssymb}
\usepackage{accessibility}
\usepackage[accsupp]{axessibility}  

\usepackage[pagebackref=true,breaklinks=true,letterpaper=true,colorlinks,bookmarks=false]{hyperref}

\newcommand{\smallsec}[1]{\vspace{0.15cm} \noindent {\bf #1.}}
\usepackage{subcaption}
\usepackage{tabularx}
\usepackage{array}
\newcommand{\PreserveBackslash}[1]{\let\temp=\\#1\let\\=\temp}
\newcolumntype{C}[1]{>{\PreserveBackslash\centering}p{#1}}
\newcolumntype{L}[1]{>{\raggedright\let\newline\\\arraybackslash\hspace{0pt}}m{#1}}
\usepackage{float}
\usepackage{placeins}
\usepackage{arydshln}

\iccvfinalcopy 

\def\iccvPaperID{7878} 

\ificcvfinal\fi

\begin{document}

\title{Understanding and Evaluating Racial Biases in Image Captioning}

\author{Dora Zhao \quad Angelina Wang \quad Olga Russakovsky\\
Princeton University \\
{\tt\small \{dorothyz,angelina.wang,olgarus\}@cs.princeton.edu}}

\maketitle
\ificcvfinal\thispagestyle{empty}\fi

\begin{abstract}
Image captioning is an important task for benchmarking visual reasoning and for enabling accessibility for people with vision impairments. However, as in many machine learning settings, social biases can influence image captioning in undesirable ways. In this work, we study bias propagation pathways within image captioning, focusing specifically on the COCO dataset. Prior work has analyzed gender bias in captions using automatically-derived gender labels; here we examine racial and intersectional biases using manual annotations. Our first contribution is in annotating the perceived gender and skin color of 28,315 of the depicted people after obtaining IRB approval. Using these annotations, we compare racial biases present in both manual and automatically-generated image captions. We demonstrate differences in caption performance, sentiment, and word choice between images of lighter versus darker-skinned people. Further, we find the magnitude of these differences to be greater in modern captioning systems compared to older ones, thus leading to concerns that without proper consideration and mitigation these differences will only become increasingly prevalent. Code and data is available at \url{https://princetonvisualai.github.io/imagecaptioning-bias/}.

\end{abstract}

\section{Introduction}
Computer vision applications have become ingrained in numerous aspects of everyday life, and problematically, so have the societal biases they contain. For example, gender and racial biases are prevalent in image tagging~\cite{schwemmer,barr2015google} and image search~\cite{kay2015unequal,noble}; visual recognition models have disparate error rates across demographics and geographic regions~\cite{pmlr-v81-buolamwini18a,devries2019everyone}. The perpetuation and amplification of social biases precipitate the need for a deeper exploration of these systems and of the bias propagation pathways. 

We focus on the task of image captioning: the process of generating a textual description of an image~\cite{vinyals2015showandtell, Luo_2018_CVPR, you2016semantic, lu2017look, anderson2018caption, huang2019attention}. 
This task serves as an important testbed for visual reasoning and can improve accessibility of digital images for people who are blind or low vision.

In this work, we assess the pathways for bias propagation: from the images, to the manual captions, and finally to the automatically generated captions. We focus our attention on studying the Common Objects in Context (COCO)~\cite{lin2014microsoft, chen2015microsoft} dataset; it is a widely used image captioning benchmark~\cite{hossain2019comprehensive}, thus making any biases especially problematic~\cite{creel2021leviathan}.
We collect both skin color and perceived gender annotations on 28,315 of the people in the COCO 2014 validation dataset after obtaining IRB approval. This data allows us (and future researchers) to analyze disparities in image captioning (and other visual recognition tasks) across different demographics. Concretely, we observe:
\begin{itemize}
\itemsep0em 
    \item The dataset is heavily skewed towards lighter-skinned (7.5x more common than darker-skinned) and male (2.0x more than female) individuals.\footnote{The gender disparity was previously observed in \cite{mals} although with automatically-inferred rather than manually-annotated labels.}  Further, darker-skinned females are especially underrepresented, appearing 23.1x less than lighter-skinned males. 
    \item There are racial terms (including racial slurs) in the manual captions. The racial descriptors are not learned by the older captioning systems~\cite{rennie2016selfcritical, Luo_2018_CVPR}, but \emph{are} learned by the newer transformer-based models~\cite{vaswani2017attention} -- although the slurs do not yet appear to be learned. 
    \item Image captioning systems perform slightly better (according to CIDEr~\cite{cider} and BLEU~\cite{papineni2002bleu}, although not SPICE~\cite{anderson2016spice}) on images of lighter-skinned people. This is consistent with disparate accuracies on e.g., pedestrian detection~\cite{wilson} and facial recognition~\cite{pmlr-v81-buolamwini18a}.   
    \item There are visual differences in the depictions of lighter and darker-skinned individuals. For example, lighter-skinned people tend to be pictured more with indoor and furniture objects, whereas darker-skinned people tend to be more with outdoor and vehicle objects.
    \item Even after controlling for visual appearance, the captions still differ in word choices used to describe images with lighter versus darker-skinned individuals.  This is particularly apparent in the manual captions and in modern transformer-based systems. 
\end{itemize}
Our work lays the foundation for studying bias propagation in image captioning on the popular COCO dataset. Data and code is freely available for research purposes at \url{https://princetonvisualai.github.io/imagecaptioning-bias/}.

\section{Related Work} 
\smallsec{Presence of dataset bias} Our work follows a long line of literature identifying, analyzing, and mitigating bias in machine learning systems. One key facet of this discussion is the bias in datasets used to train models. Under the framework of representational harms~\cite{blodgett2020nlpbias, barocas2017harms}, there is commonly a lack of representation~\cite{pmlr-v81-buolamwini18a,yang2020filter} and stereotyped portrayal~\cite{caliskan2017semantics, schwemmer,otterbacher2019nlp, miltenburg2017bias} of certain marginalized demographic groups. Along with many ethical concerns~\cite{paullada2020discontents, van2018talking}, these dataset biases are problematic because they can propagate into models~\cite{bolukbasi2016word, caliskan2017semantics}. In this work we analyze the biases present in a commonly-used image captioning benchmark, COCO~\cite{lin2014microsoft, chen2015microsoft}, using our new crowdsourced annotations.

\smallsec{Mitigating dataset bias} The root causes of dataset bias are complex: they stem from bias in image search engines~\cite{noble}, data collection practices~\cite{yang2020filter,devries2019everyone}, and real-world disparities. Proposed solutions to dataset bias include new data collection approaches~\cite{jo2020archives, hutchinson2021datasets}, manual data cleanup~\cite{yang2020filter, yang2021faceobfuscation}, synthetic data generation~\cite{ramaswamy2020debiasing,choi2020generative, sattigeri2019fairgan} -- or, in extreme cases, even withdrawing the dataset after insurmountable biases have been identified~\cite{prabhu2020large}. Researchers have advocated for increased transparency of datasets~\cite{datasheetsfordatasets, hutchinson2021datasets}, including developing tools to steer researcher intervention~\cite{AIFairness, wang2020revise}. Our work does not aim to \emph{mitigate} dataset bias but instead to articulate its impact on downstream image captioning models. 

\smallsec{Algorithmic bias mitigation}
In tandem with efforts to reform data collection, a variety of algorithmic bias mitigation techniques have been proposed; see e.g., Hutchinson and Mitchell.~\cite{mitchell2018history} for an overview. This work goes along with others that unveil biases present in existing algorithms~\cite{Obermeyer447, lum2016, angwin}. One important theme is bias amplification~\cite{mals, wang2021biasamp, wang2019balanced}, or social biases in the data getting amplified in the trained models. In this vein, we study how bias in manual image captions propagates into automated captioning systems. 

\smallsec{Image captioning models} Image captioning models are increasingly being developed as a more complex way of labeling images~\cite{vinyals2015showandtell, Luo_2018_CVPR, you2016semantic, lu2017look, anderson2018caption, huang2019attention}.
Recent work has discovered biases in these systems, but often with respect to gender~\cite{hendricks2018women, bhargava2019caption, tang2020mitigating}; the study of racial biases in captioning has been limited to analyzing bias in the \emph{manual} captions~\cite{otterbacher2019nlp, miltenburg2017bias}. Racial bias has been identified in other automated systems~\cite{benjamin2019race} (e.g., speech recognition~\cite{Koenecke7684}, facial recognition~\cite{nist}, pedestrian detection~\cite{wilson}); here we expand this work to studying racial biases in image captioning. This spurs the important question of whether race should be included in generated image captions at all. Prior works~\cite{stangl2020person, macleod2017understanding} find that, in certain contexts, people who are blind or low vision want racial descriptors to be included. Further, this motivates the need to understand how people prefer their identities labeled by an automated captioning system, a question studied extensively by Bennett et al.~\cite{bennett2021accessibility}.

\section{Crowdsourcing Demographic Annotations} 
\subsection{Annotation process}
\smallsec{Dataset} To study bias in image captioning systems, we collect annotations on COCO~\cite{lin2014microsoft, chen2015microsoft}, a large-scale dataset containing images, labels, segmentations, and 5 human-annotated captions per image. COCO is a widely used image captioning benchmark. We focus on the 40,504 images of the COCO 2014 validation set, and look for \texttt{person} instances with sufficiently large bounding boxes (at least 5,500 pixels in area) such that there is a reasonable expectation of being able to infer gender and skin color. This results in 15,762 images and 28,315 \texttt{person} instances.

\begin{figure}[t]
\begin{center}
  \includegraphics[width=\linewidth]{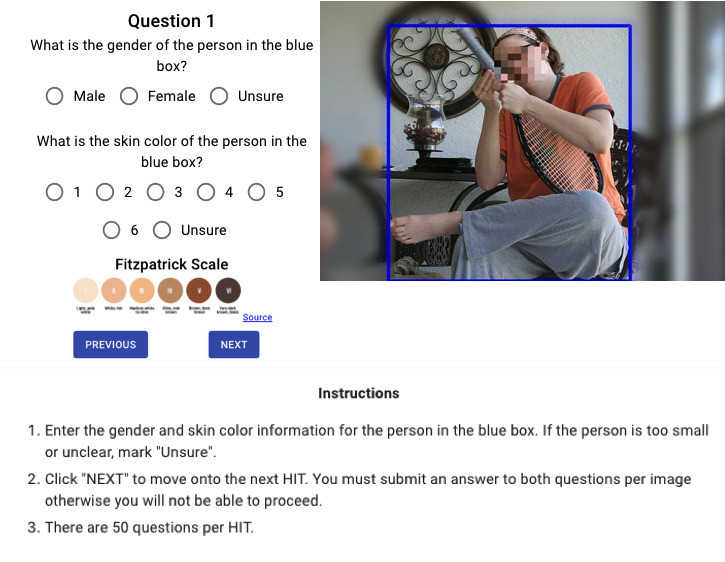}
\end{center}
  \caption{The interface shown to AMT workers, who are asked to provide the inferred gender and skin color of the un-blurred person within the blue box (pixelation not seen by annotators, only to preserve privacy in figure).}
\label{fig:amt}
\end{figure}

\smallsec{Annotation setup} Using Amazon’s Mechanical Turk (AMT), we crowdsource race and gender labels. In our interface (Fig.~\ref{fig:amt}), we present workers with a \texttt{person} instance in a COCO image and ask them to provide the skin color using the Fitzpatrick Skin Type scale~\cite{fitzpatrick1988validity}, ranging from 1 (lightest) to 6 (darkest), and the binary gender expression. We also give workers the option of marking ``unsure" for either. Each instance is annotated three times.
We compensate the workers at a rate of \$10 / hr.

\smallsec{Inferring race and gender} Race and gender annotations are fundamentally imperfect~\cite{hanna2020race, khan2021one, scheuerman2020we}. First, the annotated labels may differ from the person's identity. Second, the labels are discretized (which enables disaggregated analysis at the cost of collapsing identities). Further, the labels are for social constructs and thus subjective and influenced by the annotators' perceptions. We follow prior work~\cite{pmlr-v81-buolamwini18a,wilson} in formulating our annotation process; we use phenotypic skin color as a proxy for race because of its visual saliency over other conceptualizations of race. However, as noted by Hanna et al.~\cite{hanna2020race}, we are actualizing a particular static conceptualization of observed race here. By operationalizing race this way, we miss differences that may appear in other operationalizations, such as racial identity.

\smallsec{Quality control} To ensure annotation quality to the extent possible, we limit the task to workers who have completed over 1,000 tasks with a $98\%$ acceptance rate. We also construct 57 gold standard images where the gender and light-or-dark labels were agreed-upon by five independent annotators, including one of the authors. We inject 5 of these images randomly in a task with 50 images, and only allow workers who have correctly labeled these images to submit.

\begin{figure*}
    \centering
    \includegraphics[width=.99\linewidth]{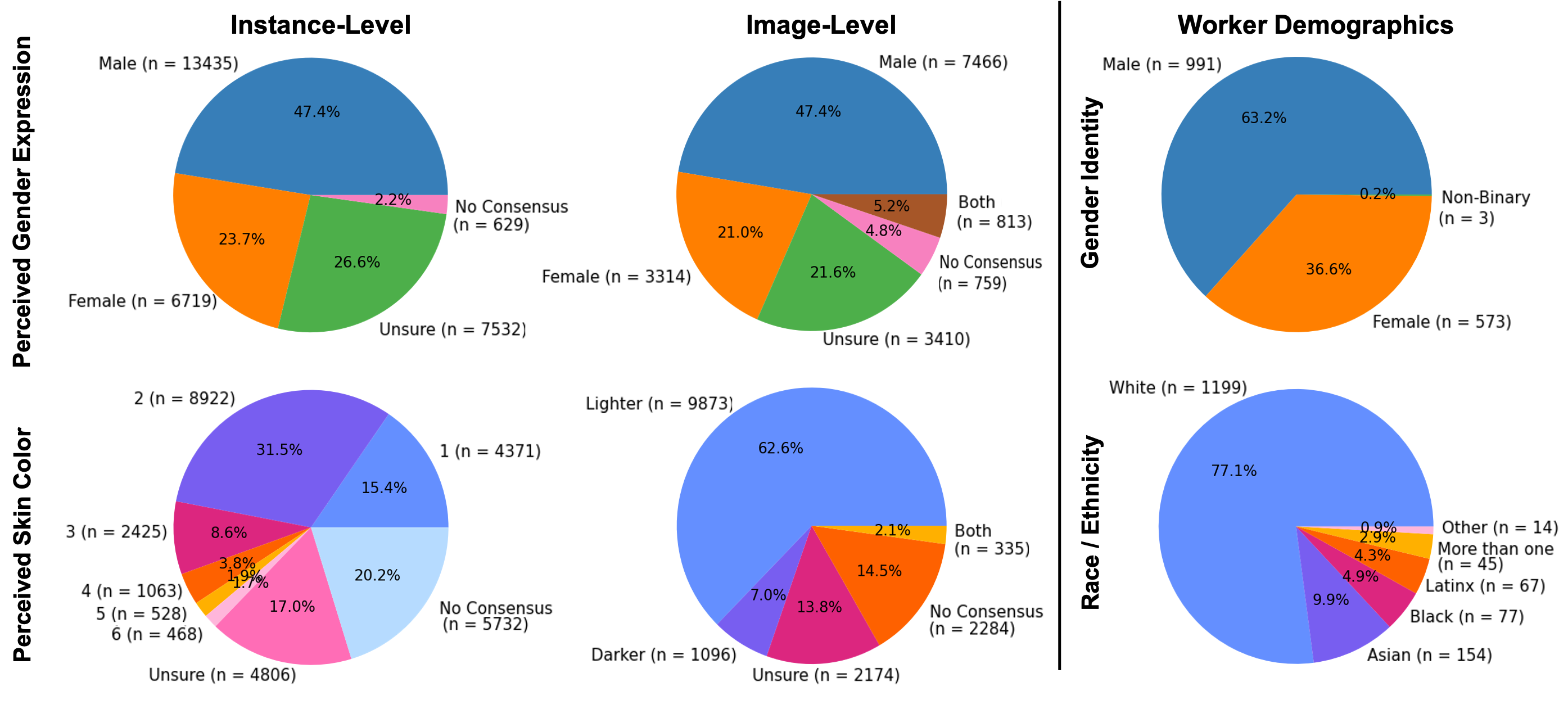}
    \caption{The results of our crowdsourced demographic annotations on the COCO 2014 validation dataset, as well as the self-disclosed demographics of the annotators. Left column: distribution of perceived skin color and gender expression of the 28,315 \texttt{people} instances. Middle column: distribution after collapsing individual annotations into image-level annotations (details in Sec.~\ref{sec:gender_annot} and~\ref{sec:skin_annot}). Lighter-skinned people and people who are male make up the majority of their respective categories. Right column: self-reported demographics of AMT workers.}
    \label{fig:amt_res}
\end{figure*}

\subsection{Gender annotations}
\label{sec:gender_annot}
We start by analyzing the collected gender annotations, looking at distributions at both the instance and image level.

\smallsec{Instance-level annotations} We analyze the gender annotations of the 28,315 \texttt{person} instances. To determine the label for a \texttt{person}, we use the majority over the three annotations. If majority is not achieved, or there are contradictory gender labels, the instance is labelled as \texttt{no consensus}. We observe that contradictory gender labels are most common when the person is a child, has obscured facial features, or possesses features that contradict social gender stereotypes (e.g. woman with short hair). 

Analyzing the distribution, we see that males make up $47.4\%$ of the instances compared to females who only comprise $23.7\%$ (see Fig.~\ref{fig:amt_res}). Most of the remaining instances were annotated \texttt{unsure} ($26.6\%$), and a consensus was unable to be reached for only $2.2\%$ of instances. 

\smallsec{Image-level annotations} To analyze the dataset at the granularity of images, which is what the captions refer to, we map individual instance annotations to the image (as there are often multiple people per image). We use the annotations given to the largest bounding box, under the assumption that captions will mainly refer to the largest person in the image~\cite{berg2012importance}. The only exception is if the second largest bounding box contains an individual of the opposite gender, and is more than half the size of the largest bounding box. In this case, we categorize the image as \texttt{both}. 

The image-level distribution closely mirrors that of the instance-level (Fig.~\ref{fig:amt_res}). Again, there are more than twice as many male images ($47.4\%$) as female images ($21.0\%$). 

\smallsec{Comparing collected gender annotations with automatically derived ones~\cite{mals}} Previously, works looking at gender bias in COCO have used gender labels derived from the manual captions: ``[if] any of the captions mention the word man or woman we mark it, removing any images that mention both genders."~\cite{mals} We compare our annotations with theirs. They label 5,413 images: our labels agree with theirs on 66.3\% and disagree on 1.4\%; the remaining 32.3\% we determine cannot be reliably labeled with one gender, e.g., because the person is too small or there are multiple people of different genders in the image. We successfully label 10,780 images; they only label 3,591 of these correctly (details in Appendix~\ref{sec:appendix_des}). This is consistent with the argument of Jacobs and Wallach~\cite{jacobs2021measurement}: gender is operationalized differently in caption-derived versus human-collected annotations.

\subsection{Skin color annotations}
\label{sec:skin_annot}
For the skin color annotations, we follow a similar process as with our gender annotations. The only difference is that we add a method for dividing skin color into the broader categories of \texttt{lighter} and \texttt{darker}. Using these new categories, we similarly analyze the skin color distribution at both the instance and image level.

\smallsec{Instance-level skin color distribution} Using the same schema as in Sec.~\ref{sec:gender_annot}, we obtain instance-level annotations for skin color. The top two most frequently occurring Fitzpatrick Skin Types are 2 ($31.5\%$) and 1 ($15.4\%$). In contrast, Fitzpatrick Skin Types 5 and 6 comprise only $1.9\%$ and $1.7\%$ of the instances, respectively. This underrepresentation of darker-skinned individuals is an example of representational harm in and of itself.

We also include a broader skin color breakdown consisting of two categories: \texttt{lighter} and \texttt{darker}. Following previous work~\cite{pmlr-v81-buolamwini18a}, we define the \texttt{lighter} category as all instances rated 1-3 on the Fitzpatrick scale and \texttt{darker} as containing 4-6. We also assign some of the instances that were previously uncategorized by skin color (because of conflicting labels assigned under the more granular 6-point scheme) to these broader categories. Using this skin color breakdown, $61.0\%$ of the instances are \texttt{lighter} individuals, whereas only $8.1\%$ are $\texttt{darker}$ individuals. The amount of \texttt{no consensus} instances decreases from $15.4\%$ to $13.9\%$ when using this breakdown.

\smallsec{Image-level skin color distribution} At the image-level, we categorize skin color as $\texttt{lighter}$ and $\texttt{darker}$, employing the same consensus method as for gender in Sec.~\ref{sec:gender_annot}. Of the images, $64.6\%$ are part of the $\texttt{lighter}$ category and $7.0\%$ are part of \texttt{darker}, meaning there are 9.2x more lighter-skinned images than darker-skinned. 

\smallsec{Intersectional analysis} We analyze the skin color and gender labels in tandem. Within $\texttt{lighter}$ images, males are overrepresented at $52.8\%$ compared to females at $25.7\%$. However, this difference is even starker when looking at $\texttt{darker}$ images, where males comprise $65.1\%$ of the images while females only make up $20.6\%$, reflecting the unique intersectional underrepresentation faced by darker-skinned females, as noted by Buolamwini and Gebru~\cite{pmlr-v81-buolamwini18a}. In fact, of the 15,762 images annotated, only 226 of them ($1.4\%$) are of darker-skinned females. 

\smallsec{Worker information} AMT workers were asked to optionally disclose their own race and gender identity. Of the workers asked, $97.9\%$ provided their gender and $97.3\%$ provided their race. As seen in Fig.~\ref{fig:amt_res}, the annotators are predominantly white ($77.1\%$) and male ($63.2\%$). 

Prior work has found that annotators describe in-group versus out-group members differently~\cite{otterbacher2019nlp}. Thus, there may be a concern that the skew in worker demographics could influence our collected labels. To understand whether a worker’s demographics influences their selection of labels, we explore disagreements in annotations. 
We do so by comparing the mean difference in annotation when the pair of workers are of the same self-reported demographic group versus when they are of differing groups. If workers from different groups label images differently, we would expect pairs from distinct groups to have a greater disagreement than pairs from the same group. However, we find for skin tone there is not a substantial difference in the disagreement between pairs of the same racial group ($0.870 \pm 0.009$) and different groups ($0.857 \pm 0.011$). For gender, the mean difference for same gender pairs ($0.109 \pm 0.002$) and different gender pairs ($0.112 \pm 0.003$) is similar as well. This indicates that there is not a systematic difference between how workers of different self-reported demographic groups label images, suggesting our collected labels would be similar even if the workers came from a different demographic composition.

\section{Experiments}
We now discuss the findings from our experiments on understanding what kinds of biases propagate in image captioning systems. First, we examine racial terms (Sec.~\ref{sec:experiments_racialterms}) and disparate performance (Sec.~\ref{sec:experiments_accuracy}). We then analyze bias in terms of representation, i.e., differences between the \texttt{lighter} and \texttt{darker} images and corresponding captions. To do this we first consider the images in Sec.~\ref{sec:experiments_visual}, before controlling for these visual differences and studying the captions in Sec.~\ref{sec:experiments_normalized}.

\smallsec{Models} We examine the captions generated by six image captioning models: (1) \textbf{FC}~\cite{rennie2016selfcritical} is a simple sequence encoder that takes in image features encoded by a CNN; (2) \textbf{Att2in}~\cite{rennie2016selfcritical} is similar but images are encoded using spatial features;  (3) \textbf{DiscCap}~\cite{Luo_2018_CVPR} further adds a loss term to encourage discriminability; (4-6) \textbf{Transformer}~\cite{vaswani2017attention}, \textbf{AoANet}~\cite{huang2019attention}, and \textbf{Oscar}~\cite{li2020oscar} are transformer-based models representing the current state-of-the-art. In our analysis we particularly focus on contrasting \textbf{Att2in} vs \textbf{DiscCap}, since they differ only in the added discriminability loss, and the older (1-3) vs the newer (4-6) models. We train the models on the COCO 2014 training set using proposed hyperparameters from the respective papers (e.g., the discriminability loss weight is $\lambda=10$ for \textbf{DiscCap}). \textbf{Oscar} is further pre-trained on a public corpus of text-image pairs 

\smallsec{Data} Our racial analysis is performed on 10,969 images of the COCO 2014 validation set which were definitively labeled as either \texttt{lighter} or \texttt{darker} (not \texttt{both} or \texttt{unsure}). 

\subsection{Captions contain racial descriptors}
\label{sec:experiments_racialterms}
We begin by analyzing the presence of racial descriptors and offensive language in the manual as well as automatically generated captions. 

\smallsec{Manual captions} Prior works~\cite{otterbacher2019nlp, miltenburg2017bias} show that people are more likely to use racial descriptors when describing non-white individuals. We observe this pattern in human-annotated captions by conducting a keyword search of the captions in the COCO 2014 training set using a precompiled list of racial descriptors (details in Appendix~\ref{sec:appendix_race}). For ambiguous terms (e.g. ``white", ``black") that can be used in a non-racial context, we manually inspect the captions. Assuming the training distribution mirrors that of the validation, for the manual captions, annotators used racial descriptors to describe individuals who appear to be white $0.03\%$ of the time versus $0.54\%$ of the time for individuals who appear to be Black. Furthermore, in $26.9\%$ of the instances when a racial descriptor for a white individual is used, the annotator is also mentioning an individual of a different race in the caption as well (e.g. ``the white woman and Black woman"). We see this as a manifestation of the belief that ``white" is the norm, and race is only salient when there is a deviation or explicit difference between multiple people.
    
In addition to looking for racial descriptors, we check for the presence of slurs and offensive language using a precompiled list of profane words~\cite{zacanger2013profane}. There are 1,691 instances of profane language, occurring in $0.40\%$ of the sentences in the COCO 2014 training set. We find alarming occurrences not only of racial slurs but also of homophobic and sexist language as well, similar to the NSFW discoveries by Prabhu and Birhane~\cite{prabhu2020large}.

\smallsec{Automated captions}
Racial descriptors are not found in the automated captions generated by $\textbf{FC}$, $\textbf{Att2In}$, $\textbf{DiscCap}$, \textbf{AoANet}, or \textbf{Oscar}. While this may be attributed to the fact that racial descriptors are uncommon in the training set, we disprove the idea that this is wholly the reason. To do so, we observe that other words which occur at similar rates (and are thus equally uncommon) are in fact still present in the model-generated captions. For example, the word ``Japanese" occurs 69 times in the training set and 0 times in \textbf{AoANet}-generated captions while other descriptors, such as ``uncooked" and ``soaked", which appear 88 and 61 times in the training set, occur 2 and 6 times in the generated captions respectively.
 
\begin{figure}
    \centering
    \includegraphics[width=8.5cm]{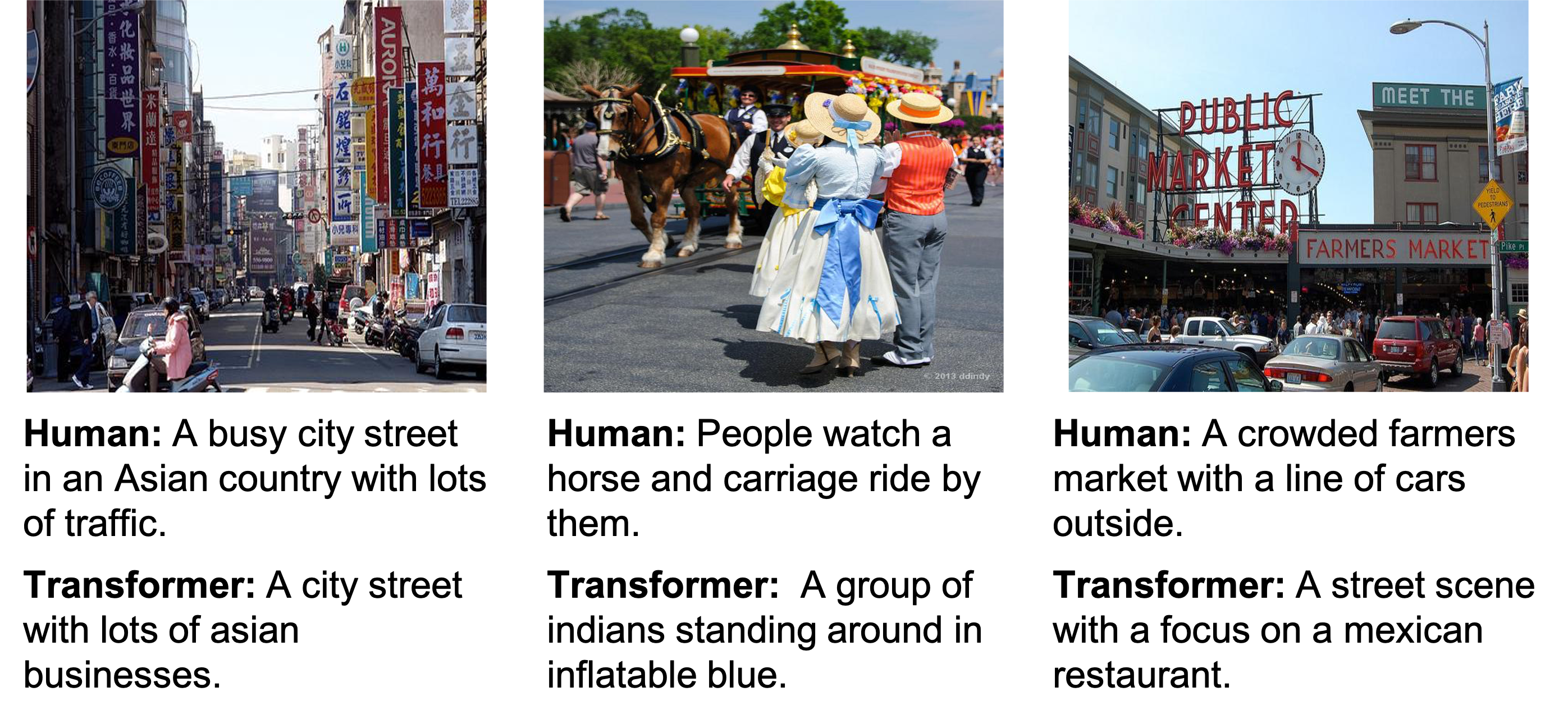}
    \caption{Examples of images for which the \textbf{Transformer} model~\cite{vaswani2017attention} assigns racial or cultural descriptors to the caption. While in the first image the descriptor of ``Asian" is present in the human-annotated caption, neither of the descriptors, ``Indian" nor ``Mexican," are applicable in the latter images.}
    \label{fig:transf_qual}
\end{figure}

While it is rare, we find that racial and cultural descriptors as well as offensive language do propagate into the captions generated by the newer transformer-based models. For \textbf{Transformer}, \textbf{AoANet}, and \textbf{Oscar}, we find instances of offensive language. In addition, there are racial descriptors in 2 of the captions generated by \textbf{Transformer} and 12 cultural descriptors. Furthermore, for 10 of the 14 images, the model uses these descriptors when the human captions do not contain any racial or cultural descriptors (Fig.~\ref{fig:transf_qual}). This leads to the worry that models may replicate offensive language or exploit spurious correlations to assign descriptors in a stereotypical and harmful way.

\subsection{Performance differs slightly between \textbf{\texttt{lighter}} and \textbf{\texttt{darker}} images}
\label{sec:experiments_accuracy}

We next evaluate whether image captioning models produce captions of different qualities on images with lighter-skinned people than darker-skinned people. To do so, we first assess the differences in BLEU~\cite{papineni2002bleu}, CIDEr~\cite{cider} and SPICE~\cite{anderson2016spice} scores between captions on \texttt{lighter} and \texttt{darker} images. Both BLEU and CIDEr rely on \textit{n}-gram matching with BLEU measuring precision and CIDEr the similarity between the generated caption and the ``consensus" of manual captions. SPICE, however, focuses more on semantics, capturing how accurately a generated caption describes the image's scene graph (e.g. objects, attributes).

\begin{figure}
    \centering
    \includegraphics[width=8cm]{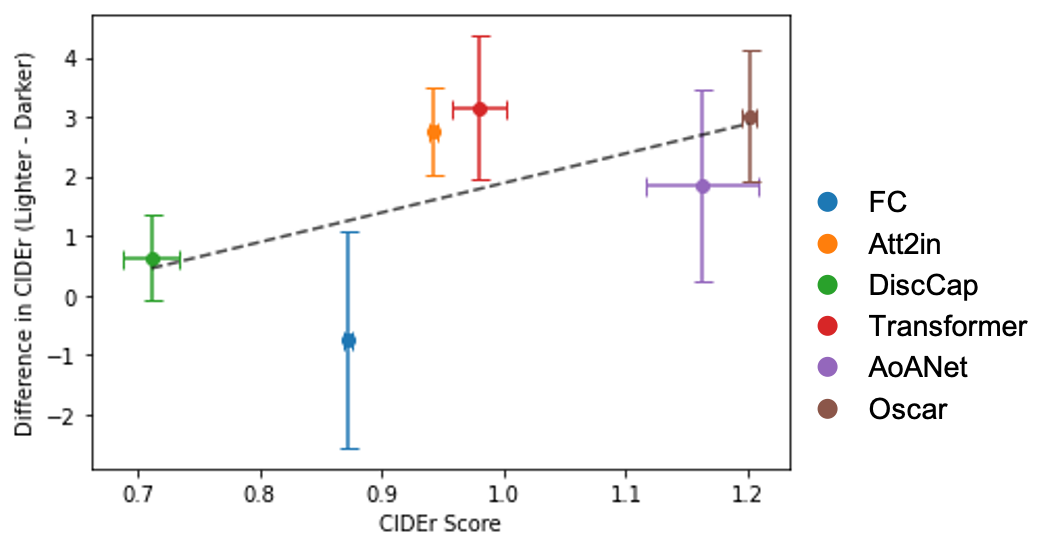}
    \caption{Regressing model performance, as measured by CIDEr~\cite{cider}, against difference in performance (CIDEr on \texttt{lighter} - CIDEr on \texttt{darker}) suggests that as performance increases, the difference may correspondingly increase as well ($R^2= 0.343$). The horizontal and vertical error bars in the graph represent the $95$\% confidence intervals for the performance and differences, respectively}
    \label{fig:ciderdelta}
\end{figure}

From these results (Tbl.~\ref{tbl:comparison_acc}), we make two key observations. First, according to both BLEU and CIDEr, the models \textbf{Att2in}, \textbf{Transformer}, \textbf{AoANet}, and \textbf{Oscar} perform somewhat better on \texttt{lighter} images than \texttt{darker} images: e.g., they achieve $2.7\pm 0.7$, $3.2\pm1.2$, $1.9 \pm 1.6$, and $3.0 \pm 1.1$ higher CIDEr scores respectively on \texttt{lighter} than \texttt{darker} images. We observe that these differences in BLEU and CIDEr are not significant for the \textbf{FC} and \textbf{DiscCap} --- likely because their overall CIDEr scores are worse, at only  $87.2$ and $71.1$ respectively, whereas the other four models attain CIDEr scores above $90.0$ (see Appendix~\ref{sec:appendix_performance}). This suggests that the \emph{way} models are choosing to describe the images may be better-suited for the majority group. In fact, we see there is a slight positive correlation between the performance of the model (as measured by CIDEr) and the differences in performance between the two groups with an $R^2$ of $0.343$ (Fig.~\ref{fig:ciderdelta}). Second, there are no noticeable differences with SPICE, indicating that the captions identify key visual concepts equally accurately across both groups. Nonetheless, it is important to note that negative results do not indicate something is bias-free, but merely that our particular experiment did not uncover strong biases.

\begin{table}[t]
\begin{center}
\caption{The differences in captioning performance (score on \texttt{lighter} - score on \texttt{darker}) as measured by BLEU~\cite{papineni2002bleu}, CIDEr~\cite{cider}, and SPICE~\cite{anderson2016spice} multiplied by 100 on the COCO 2014 validation dataset. Error bars represent $95\%$ confidence intervals across random seeds used to train 5 models per architecture.}
\label{tbl:comparison_acc}
\begin{tabular}{l|ccc}
\hline
& BLEU & CIDEr &  SPICE\\
\hline
FC~\cite{rennie2016selfcritical} &  $0.5\pm0.5$& $-0.8 \pm 1.8$ & $0.2 \pm 0.3$\\
Att2in~\cite{rennie2016selfcritical} & $2.4\pm0.4$  &$2.7 \pm 0.7$ & $0.0 \pm 0.1$\\
DiscCap~\cite{Luo_2018_CVPR} & $0.3\pm0.5$ & $0.6 \pm 0.7$& $0.0 \pm 0.2$\\
\hdashline
Transformer~\cite{vaswani2017attention} & $2.5\pm0.9$ &  $3.2 \pm 1.2$ &$-0.1 \pm 0.3$\\
AoANet~\cite{huang2019attention} & $1.8 \pm 0.8$ & $1.9 \pm 1.6$ & $0.0 \pm 0.2$ \\
Oscar~\cite{li2020oscar} & $2.1 \pm 0.7$ & $3.0 \pm 1.1$ & $0.1 \pm 0.3$
\end{tabular}
\end{center}
\end{table}

\subsection{Visual appearance differs between \textbf{\texttt{lighter}} and \textbf{\texttt{darker}} images}
\label{sec:experiments_visual}
The analyses so far only consider issues in the captions themselves, irrespective of the image. We now explore how the visual depictions of people of different groups differ. We analyze simple image layout statistics, apply the REVISE~\cite{wang2020revise} tool for discovering bias in datasets, and consider differences in visual appearance of the image content. 

We split our skin-tone-labeled image dataset of 10,969 images into 9,609 images for training and 1,360 for testing.\footnote{These images belong to the COCO 2017 training and validation set respectively; recall that all belong to the COCO 2014 validation set.} We use area under the ROC curve (AUC) as our metric on a balanced (through re-weighting) test set, so random guessing would have an AUC of $50\%$. We bootstrap over 1,000 resamples and report a $95\%$ confidence interval. 

\smallsec{Image layout statistics} We consider the following simple image layout statistics as our features: number of people in the image, largest person bounding box size, distance of the largest bounding box from the center of the image, and gender (\texttt{male}, \texttt{female}, \texttt{unsure}, or \texttt{no consensus}) one-hot coded. We train logistic regression models using LBFGS through the \texttt{sklearn} package~\cite{scikit} to predict whether the input corresponds to the \texttt{lighter} or \texttt{darker} label. An ability to classify serves as a signal for how distinguishable the input features of the two groups are. We use a balanced class weight and run five-fold cross-validation to tune the L2 regularization hyperparameter ($1\mathrm{e}{-4}$ to $1\mathrm{e}{4}$). 

Our two best performing models are trained on the distance from center and the distance plus the gender. Distance alone achieves an AUC of $56.6 \pm 5.2$; adding gender increases the AUC to $57.8 \pm 4.9$. Distance is predictive because darker-skinned individuals tend to be further from the image center than lighter-skinned individuals; this is troubling since the ``important" parts of an image tend to be more centered~\cite{berg2012importance}. 
Gender is a useful feature since from Sec.~\ref{sec:skin_annot} we know that the gender distribution differs between the two groups. 

\begin{figure}
    \centering
    \includegraphics[width=7.5cm]{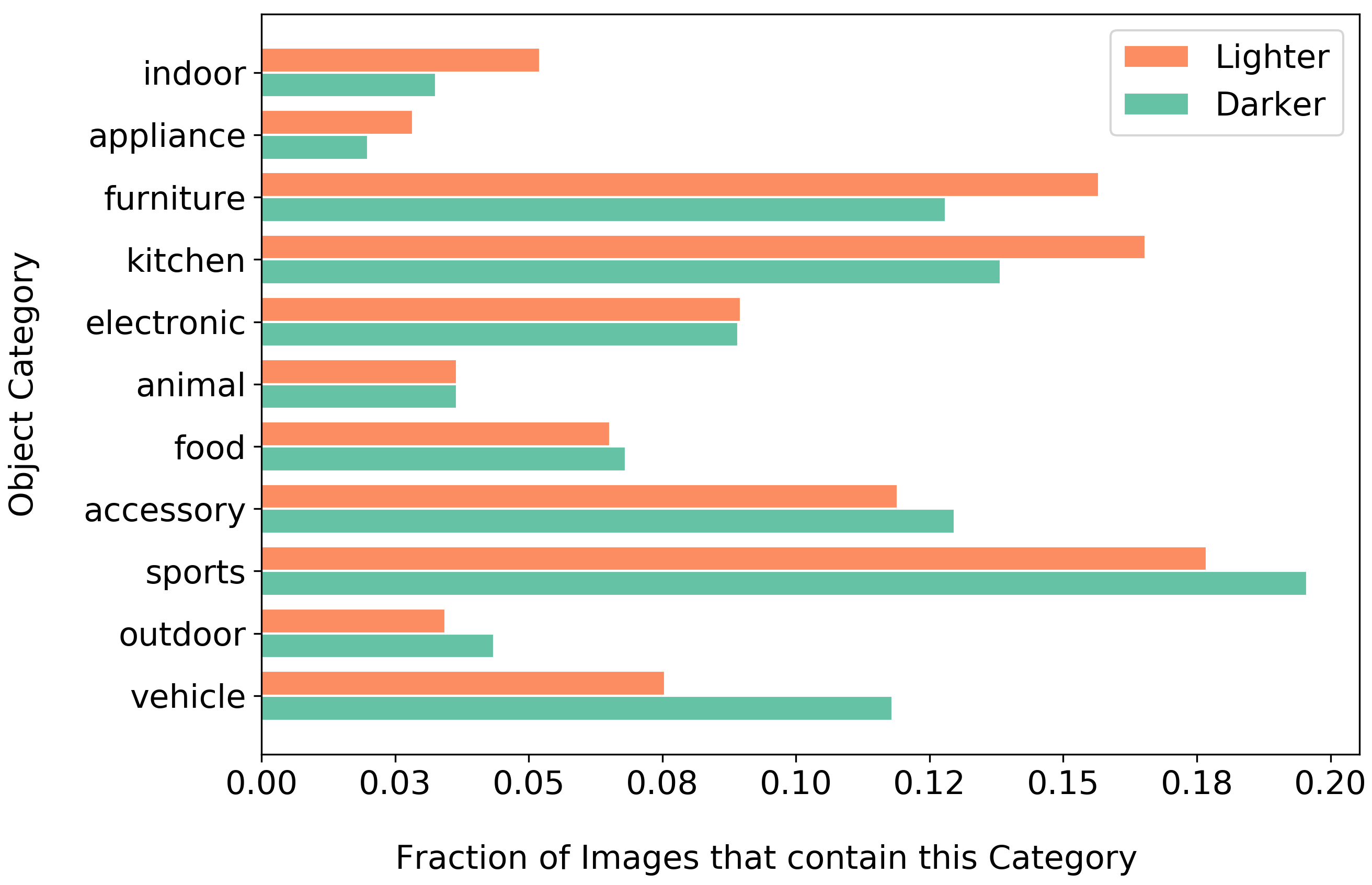}
    \caption{Images with people of lighter and darker skin tones co-occur with object categories at different frequencies. Whereas the former tend to be pictured with object categories that are indoor, the latter tend to be pictured with object categories that are more likely to be outdoors.}
    \label{fig:revise_objcat}
\end{figure}

\smallsec{REVISE~\cite{wang2020revise} bias discovery} We next apply the REvealing VIsual biaSEs (REVISE) tool.\footnote{We additionally include the 813 images labeled \texttt{both} in both groups.}
Using REVISE we discovered that darker-skinned people appear more frequently with outdoor objects, and lighter-skinned people appear more frequently with indoor objects (Fig.~\ref{fig:revise_objcat}). Specifically, objects like sink, potted plant, and toothbrush all appear with lighter-skinned people over 13x as much as with darker-skinned people, despite lighter-skinned people only appearing in 7x as many images as darker-skinned people. Although at the moment the differences in object co-occurrences do not appear to have noticeable downstream effects (Sec.~\ref{sec:experiments_accuracy}), these differences may lead to discrepancies in performance as certain objects become more easily identifiable for different skin tone groups. 

\smallsec{Visual appearance} Finally, we use image classification models for a detailed examination of how the content of the images differs between different skin tones. To ensure that the skin color of the pictured individual does not affect the model's prediction, we use COCO's object-level segmentations to mask all the \texttt{people} objects. We fill in these masks with the average color pixel in the image. Using the masked images, we fine-tune a pre-trained ResNet-101~\cite{resnet} over five epochs using the Adam optimizer~\cite{kingma2015adam} and a batch size of 64. We oversample the \texttt{darker} images to account for the imbalanced class sizes. During training, the learning rate is initialized to be 0.01 and decays by a factor of 0.1 after three epochs. The model achieves an AUC of $55.4 \pm 4.9$, indicating that there is a slight learnable difference between the scenes of \texttt{lighter} and \texttt{darker} images. 

\begin{table}[t]
\begin{center}
\caption{Three bias analyses on manual and automated captions of images for which visual content has been controlled. For the first column the VADER sentiment score~\cite{hutto2014vader} is multiplied by 100. For the last two columns, the number is AUC$\times$100 for classification ability, where higher numbers indicate a greater ability to distinguish between the two groups. Error bars represent $95\%$ confidence intervals across random seeds used to train 5 models per architecture.}
\label{tbl:comparison_normalized}
\begin{tabular}{l|c|cc}
\hline
& Sentiment  & Embedding & Vocab. \\
&($\Delta$)& (AUC) & (AUC) \\
\hline
Human  & $1.5$ & $68.9$  & $61.8$\\
\hdashline 
FC~\cite{rennie2016selfcritical} &$1.0 \pm 0.4$ &  $55.8 \pm 5.7$ & $65.9 \pm 4.2$\\
Att2in~\cite{rennie2016selfcritical}&  $0.3 \pm 0.2$ &  $55.3 \pm 2.5$ & $62.8 \pm 1.5$\\
DiscCap~\cite{Luo_2018_CVPR}& $0.9 \pm 0.7$ &  $52.2 \pm 2.8$ & $63.0 \pm 3.2$\\
\hdashline 
Transformer~\cite{vaswani2017attention} & $0.6 \pm 0.8 $ & $54.2\pm 3.1$ & $66.0 \pm 1.4$\\
AoANet~\cite{huang2019attention} & $1.0 \pm 0.4$ & $56.3 \pm 3.0$ & $68.0 \pm 1.8$ \\
Oscar~\cite{li2020oscar} & $0.6 \pm 0.6$ & $54.0 \pm 2.6$ & $64.4 \pm 2.8$
\end{tabular}
\end{center}
\end{table}

\subsection{Captions describe people differently based on skin tone}
\label{sec:experiments_normalized}
\begin{figure}
\centering
\includegraphics[width=.85\linewidth]{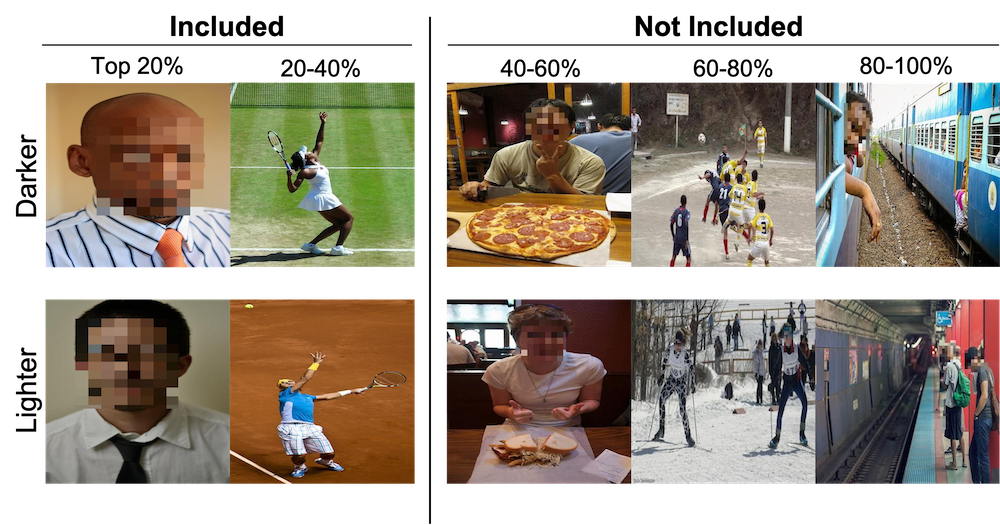}
\caption{Examples of paired images along 20 percentile increments of similarity. The leftmost images represent an example pair from the most similar top $20\%$ of pairs, and the rightmost represent the bottom $20\%$. We pick $40\%$ as our threshold for controlled images to include.}
\label{fig:simimage}
\end{figure}

Finally, we consider how both manual and automatic captions differ when describing \texttt{lighter} versus \texttt{darker} images. To do so, we first control for the visual differences, in order to disentangle the issues coming from the image content versus from the words used in the caption. We do so by finding images that are as similar as possible in content, and differ only by the skin color of the people pictured, i.e., constructing counterfactuals within the realm of our existing dataset. Concretely, for each \texttt{darker} image, we find the corresponding \texttt{lighter} image that minimizes the Euclidean distance between the extracted ResNet-34 features~\cite{resnet} of the masked images using the Gale-Shapley algorithm~\cite{gale1962college} for stable matching (Fig.~\ref{fig:simimage}). After examining the results, we select the top $40\%$ most similar image pairs.

The resulting dataset has 876 images. When needed, we use 700 for training (80\%) and 176 for testing (20\%); otherwise we compute statistics over the whole dataset. As expected, a visual classifier trained on these images (with the people masked) achieves an AUC of only $44.7 \pm 9.3$, failing to differentiate between the two groups. 

In the following analyses, we use the same six models and training setup as in previous experiments. However, we use the dataset, introduced above, which consists of 876 unmasked images for evaluation. This data thus allows us to examine whether human-annotated and model-generated captions diverge even when visual differences (except skin color) are controlled.

\subsubsection{Sentiment Analysis}
For our first line of inquiry, we use the Valence Aware Dictionary and Sentiment Reasoner (VADER)~\cite{hutto2014vader} to perform sentiment analysis on the human-annotated captions. Limitations include that sentiment analysis tools have been shown to encode societal biases themselves~\cite{kiritchenko2018sentiment, davidson2019bias}, and may not generalize well to out-of-distribution machine-generated text. VADER returns a compound polarity score from $-1$ (strongly negative) to $1$ (strongly positive). Scores less than $-0.05$ are considered negative; scores greater than $0.05$ positive. We find that human-annotated captions describing \texttt{lighter} images
have a mean compound score of $0.073 \pm 0.01$ whereas those describing \texttt{darker} images have a mean compound score of $0.059 \pm 0.01$. The difference in compound scores is statistically significant ($p = 0.005$), with captions describing \texttt{lighter} images being more positive. 

We find that automated captioning systems do not appear to amplify the difference in sentiment scores between the two groups (Tbl.~\ref{tbl:comparison_normalized}). The lack of difference is largely due to the fact that automated captions tend to be more neutral than the human-annotated ones, thus removing most of the sentiment. In fact, the compound scores were all less than 0.03, excluding scores for captions generated by \textbf{Transformer} (0.046 for \texttt{lighter} and 0.042 for \texttt{darker}).

\subsubsection{Sentence embedding differences} For our next analysis, we use sentence embeddings from the Universal Sentence Encoder~\cite{cer2018universal} to compare how the semantic content of captions differs between \texttt{lighter} and \texttt{darker} images. To note, racial descriptors in the captions are not removed for this experiment.

We train a multilayer perceptron classifier (MLP) on the embeddings and run five-fold cross validation to tune the learning rate ($1\mathrm{e}{-5}$ to $1$) and number of epochs (1 to 150). 

We find that the classifier can differentiate between the captions with an AUC of $68.9 \pm 3.5$, indicating a learnable difference in the resulting caption content despite the visual content (with skin tone masked) being indistinguishable.

We see in Tbl.~\ref{tbl:comparison_normalized} that the ability to differentiate based on embeddings drops in the generated captions, especially for the more advanced \textbf{Transformer} model to $54.2 \pm 3.1$, which is almost random. Although humans appear to be assigning different content to similar images with people of different skin tones, automated captioning models do not appear to uphold this trend, at least with respect to the particular sentence embeddings we use.
    
\subsubsection{Vocabulary differences}
Finally, we consider word choice in the captions. We use a logistic regression model and a vocabulary of the 100 most commonly used words (filtering out articles, prepositions, and racial descriptors, e.g. ``white") in the COCO 2014 training set. Our features are size 100 binary indicators of whether a particular word is present in a caption. The classifier achieves an AUC of $61.8 \pm 3.8$ on human captions. Beyond the differential use of racial descriptors we already observed in Sec.~\ref{sec:experiments_racialterms}, this suggests annotators use different vocabularies to describe images even with similar visual content (other than skin tone). 

The ability to distinguish between \texttt{lighter} and \texttt{darker} images further increases when automated captions are used. Particularly, in Tbl.~\ref{tbl:comparison_normalized} we see from \textbf{Att2in} to \textbf{DiscCap} and \textbf{FC} to \textbf{Transformer}, the AUCs slightly increases from $62.8\pm 1.5$ to $63.0\pm3.2$ and $65.9\pm 4.2$ to $66.0 \pm 1.4$, respectively. From \textbf{FC} to \textbf{AoANet}, there is a greater increase in AUC from $65.9\pm 4.2$ to $68.0 \pm 1.8$. We do note that, for \textbf{Oscar}, the ability to differentiate based on vocabulary decreases compared to \textbf{FC} as the AUC drops from $65.9 \pm 4.2$ to $64.4\pm 2.8$. This may be due to the fact that \textbf{Oscar} is pre-trained on a larger corpus of data; the greater dataset diversity may help diminish the differences between the vocabularies used. Overall, this leads us to believe that more advanced models are more likely to employ different word choices when describing different groups of people.

Interpreting these results relative to that of the previous section in which we found that the semantic content of generated captions did not differ much between different groups, we consider whether different words are being used despite caption content being similar. As an example, the sentences ``Apples are good." and ``Apples are great." may map to similar sentence embeddings, but the specific word choice employed is different. In this vein, we find, for instance, that on \textbf{AoANet}'s captions, the average coefficent of the word ``road" is $0.226$ higher than that of the word ``street" (where higher coefficients are predictive of \texttt{darker}), even though upon manual inspection the images being described are similar (see Appendix~\ref{sec:appendix_vocab}). While differences in the usage of words, such as ``road" and ``street," are relatively innocuous, these subtle differences in vocabulary may become more problematic when we consider how certain words like ``articulate" have developed a different meaning when applied to Black people~\cite{clemetson2017speaking, alim2012articulate}. Thus, in future work, it is important to consider not only the semantic differences captured in the sentence embeddings but also the specific words being employed.

\section{Discussion and Conclusion}
In this work, we seek to understand not only what racial biases are present in the COCO image captioning dataset, but also how these biases propagates into models trained on them. We annotate skin color and gender expression of people in the images, and consider various forms of bias such as those in the form of differentiability between different groups. We find instances of bias in the dataset and the automated image captioning models. However, we are careful to note that cases in which we did not find bias do not mean there are not any, merely that our particular experiments did not uncover them. By looking at the models that seem to be most indicative of where the image captioning space is progressing, we can see that the bias appears to be increasing. For researchers, this serves as a reminder to be cognizant that these biases already exist and a warning to be careful about the increasing bias that is likely to come with advancements in image captioning technology.

Based on these analyses, we propose directions for mitigating the biases found in captioning systems. First, from our findings in Sec.~\ref{sec:gender_annot} and~\ref{sec:experiments_racialterms}, we see that human annotators make assumptions about the demographics of people pictured or use different language when describing people of different skin tone groups. To mitigate this, dataset collectors can provide more explicit instructions for annotators (e.g. do not label gender or include racial descriptors to people). In addition, we also find that ground-truth captions contain profane language (Sec.~\ref{sec:experiments_racialterms}). In line with existing mitigation efforts~\cite{yang2020filter, prabhu2020large}, manual captions containing slurs or other offensive concepts should be removed from the dataset. Additionally, in Fig.~\ref{fig:amt_res} we see that only $7.0\%$ of the dataset contained images of people with darker skin tones, i.e., 1096 images. We need to collect more diverse datasets such that we can measure disaggregated statistics and compare metrics such as the difference in SPICE scores with the knowledge that our measurements do not suffer from a high sampling bias. Finally, from our analysis of generated captions (Sec.~\ref{sec:experiments_normalized}), we note that \textbf{Oscar} exhibits less bias compared to the other transformer-based models. This suggests the greater dataset diversity from pre-training the model may help reduce the amount of bias that propagates into the automated captions.

\begin{flushleft}
\smallsec{Acknowledgements} This work is supported by the National Science Foundation under Grant No. 1763642 and the Friedland Independent Work Fund from Princeton University's School of Engineering and Applied Sciences. We thank Arvind Narayanan, Karthik Narasimhan, Sunnie S. Y. Kim, Vikram V. Ramaswamy, and Zeyu Wang for their helpful comments and suggestions, as well as the Amazon Mechanical Turk workers for the annotations.
\end{flushleft}
{\small
\bibliographystyle{ieee_fullname}
\bibliography{egbib,visualai}
}

\appendix
\begin{flushleft}
\textbf{{\huge Appendix}}
\end{flushleft}
\section{Comparing collected gender annotations with automatically derived ones} 
\label{sec:appendix_des}
We explore extending the schema introduced in Sec.~\ref{sec:gender_annot} for deriving gender labels from captions in three ways: 1) only labeling images where there is a \texttt{person} who has a bounding box greater than 5,500 pixels, 2) expanding the list of gendered words beyond ``man" and ``woman", and 3) having different cutoffs for how many captions (of the 5 per image) need to mention a gender for the image to be labeled. We call the use of the gendered set $\{$man, woman$\}$ ``few," and that of our expanded set ``many."

Our expanded set ``many" consists of the following words: [``male", ``boy", ``man", ``gentleman", ``boys", ``men", ``males", ``gentlemen"] and [``female", ``girl", ``woman", ``lady", ``girls", ``women", ``females", ``ladies"].

\begin{figure}[htbp]
\centering
\begin{subfigure}{.48\textwidth}
  \centering
  \includegraphics[width=.98\linewidth]{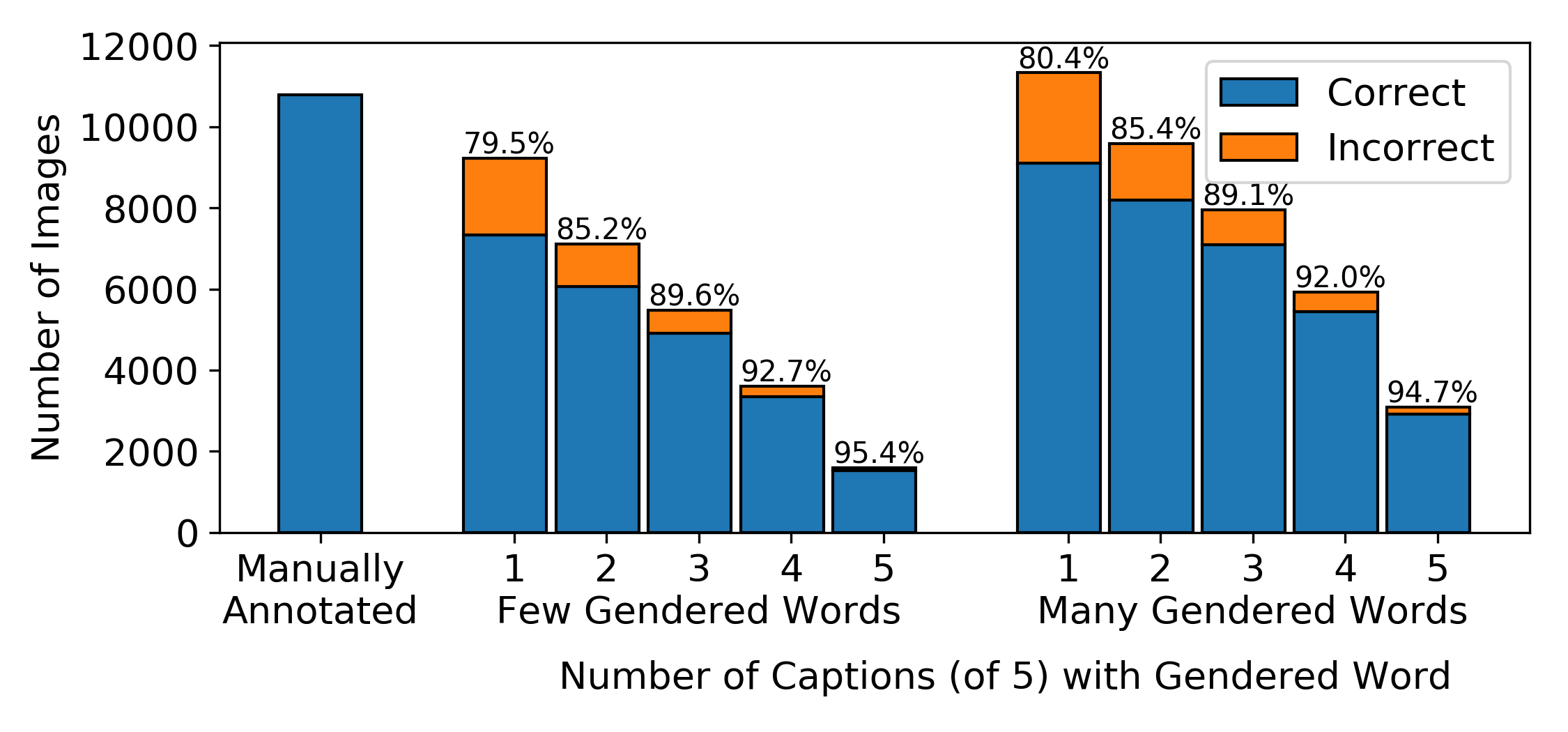}
\end{subfigure}
\caption{Comparison of various ways of automatically deriving image-level gender annotations from existing image captions. ``Few" and ``Many" gendered words refers to the size of the set of gendered words considered, and numbers refer to required captions that mention a gendered word, higher numbers limiting the images that can be labeled. Percentage over each bar indicates accuracy. All methods are imperfect and noisy, cautioning against the use of automatically-derived gender annotations.}
\label{fig:gen_comp}
\end{figure}

Our results in Fig.~\ref{fig:gen_comp} show that while these extensions significantly increase both the number of images correctly labeled and the accuracy of labeled images, all methods are inaccurate and/or incomplete. 
The gender labels derived from captions remain highly imperfect, as expected, cautioning against automated means of gender derivation~\cite{keyes2018agr}. 

\section{Racial descriptors}
\label{sec:appendix_race}
When searching for descriptors of race and ethnicity in Sec.~\ref{sec:experiments_racialterms}, we first convert the captions to lowercase. We then use the following keywords [``white", ``Caucasian", ``Black", ``African", ``Asian", ``Latino", ``Latina", ``Latinx", ``Hispanic", ``Native", and ``Indigenous"] --- also in lowercase --- to query the captions. 

\section{Caption performance} 
\label{sec:appendix_performance}
In Sec.~\ref{sec:experiments_accuracy} we assess the differences in caption performance for BLEU~\cite{papineni2002bleu}, CIDER~\cite{cider}, and SPICE~\cite{anderson2016spice} when evaluated on the COCO 2014 validation set. We extend this analysis by providing the overall scores across the four image captioning models and looking at two additional automated image captioning metrics. 

To start, we look at the performance for our four models. As seen in Tbl.~\ref{tbl:allscores}, \textbf{Oscar} has the best performance across all metrics. Further, we see that the newer transformer-based models outperform older models (e.g. \textbf{FC}, \textbf{Att2in}, and \textbf{DiscCap}) across all metrics as well. 

We also report the differences in performance between \texttt{lighter} and \texttt{darker} images for two commonly used image captioning metrics --- METEOR~\cite{banerjee2005meteor} and ROUGE~\cite{lin2004rouge} (Tbl.~\ref{tbl:alldelta}). Similar to the results for BLEU and CIDER, the differences for METEOR and ROUGE are greater for \textbf{Att2in}, \textbf{Transformer}, and \textbf{Oscar}. \textbf{AoANet} also shows some slight differences in performance for METEOR and ROUGE. This supports our observation that the better performing captioning models also tend to show greater discrepancies in performance between \texttt{lighter} and \texttt{darker} images.

\begin{table}[htbp]
\begin{center}
\caption{The caption performance as measured by BLEU~\cite{papineni2002bleu}, CIDEr~\cite{cider}, and SPICE~\cite{anderson2016spice} multiplied by 100 on the COCO 2014 validation dataset. Error bars represents $95\%$ confidence intervals across random seeds used to train 5 models per architecture.} 
\label{tbl:allscores}
\begin{tabular}{l|ccc}
\hline
& BLEU & CIDEr & SPICE \\ 
\hline 
FC~\cite{rennie2016selfcritical} & $28.2 \pm 0.4$ & $87.2 \pm 0.4$ & $16.9 \pm 0.2$\\ 
Att2in~\cite{rennie2016selfcritical} & $31.5 \pm 0.2$ & $94.2 \pm 0.3$ & $18.4 \pm 0.1$\\ 
DiscCap~\cite{Luo_2018_CVPR} & $23.7 \pm 0.7$ & $71.1 \pm 2.3$ & $18.8 \pm 0.2$\\
\hdashline
Transformer~\cite{vaswani2017attention} & $33.9 \pm 0.6$ & $97.2 \pm 2.2$  & $20.2 \pm 0.2$\\
AoANet~\cite{huang2019attention} & $38.7\pm1.6$ & $116.2\pm4.6$ & $22.2\pm 0.6$\\
Oscar~\cite{li2020oscar} & $40.0 \pm 0.4$ & $120.0 \pm 0.5$ & $23.0 \pm 0.3$\\
\end{tabular}
\end{center}
\end{table}

\begin{table*}[!h]
\begin{center}
\caption{The  differences  in  captioning  performance (score on \texttt{lighter}- score on \texttt{darker}) as measured by BLEU~\cite{papineni2002bleu}, METEOR~\cite{banerjee2005meteor}, ROUGE,~\cite{lin2004rouge}, CIDEr~\cite{cider}, and SPICE~\cite{anderson2016spice} multiplied by 100 on the COCO 2014 validation dataset. Error bars represent 95\% confidence intervals across random seeds used to train 5 models per architecture}
\label{tbl:alldelta}
\begin{tabular}{l|ccccc}
\hline
& BLEU $\Delta$ & METEOR $\Delta$ & ROUGE $\Delta$ & CIDEr $\Delta$ & SPICE $\Delta$ \\
\hline
FC~\cite{rennie2016selfcritical} & $0.5 \pm 0.5$ & $-0.1 \pm 0.2$ & $0.0 \pm 0.5$ &  $-0.8 \pm 1.8$ & $0.2 \pm 0.3$\\
Att2in~\cite{rennie2016selfcritical} & $2.4 \pm 0.4$ & $0.5 \pm 0.1$ & $0.8 \pm 0.2$ & $2.7 \pm 0.7$ & $0.0 \pm 0.1$ \\
DiscCap~\cite{Luo_2018_CVPR} & $0.3 \pm 0.5$ & $0.0 \pm 0.3$ & $0.0 \pm 0.3$ & $0.6 \pm 0.7$ & $0.0 \pm 0.2$\\
\hdashline 
Transformer~\cite{vaswani2017attention} & $2.5 \pm 0.9$ & $0.6 \pm 0.3 $ & $1.1 \pm 0.4$ & $3.2\pm 1.2$ & $-0.1 \pm 0.3$\\
AoANet~\cite{huang2019attention} & $1.8 \pm 0.8$ & $0.3 \pm 0.3$ & $0.4 \pm 0.3$ &$1.9\pm 1.6$ & $0.0 \pm 0.2$\\
Oscar~\cite{li2020oscar} & $3.0 \pm 1.1$ & $0.5 \pm 0.4$ & $0.8 \pm 0.3$ & $3.0 \pm 1.1$ & $0.1 \pm 0.3$
\end{tabular}
\end{center}
\end{table*}

\begin{table*}[!htb]
    \setlength{\tabcolsep}{3pt}
    \centering
    \caption{The most predictive words of $\texttt{lighter}$ and $\texttt{darker}$ images for a logistic regression trained on manual and automated captions of images for which visual content has been controlled. Lower coefficients are more predictive of \texttt{lighter}. For the automated captions, the coefficients are averaged across the 5 models per architecture.}
    \begin{tabular}{ll|ll|ll|ll|ll|ll|ll}
        \hline 
        \multicolumn{2}{{l|}}{Human} & \multicolumn{2}{{l|}}{FC~\cite{rennie2016selfcritical}} & 
        \multicolumn{2}{{l|}}{Att2in~\cite{rennie2016selfcritical}} & 
        \multicolumn{2}{{l|}}{DiscCap~\cite{Luo_2018_CVPR}} & 
        \multicolumn{2}{{l|}}{Transformer~\cite{vaswani2017attention}} & 
        \multicolumn{2}{{l|}}{AoANet~\cite{huang2019attention}} & 
        \multicolumn{2}{{l|}}{Oscar~\cite{li2020oscar}}
        \\
        \hline
         Train & -0.13 & Parked & -0.20 & Table & -0.28 & Tree & -0.26 & Tree & -0.24 & Bathroom & -0.22 & Brown & -0.21\\
         Cake & -0.12 & Young &-0.15 & Tree & -0.22 & People & -0.17 & Snow &-0.17 & Tree & -0.19 & Baseball & -0.18\\
         Man & -0.11 & Holding &-0.15 & City & -0.22 & Brown & -0.17 & Two &-0.15 & Baseball & -0.18 & Bathroom & -0.16\\
         Playing & -0.10 & Wearing& -0.14 & Field & -0.19 & Two & -0.15 & Bear & -0.14 & Brown & -0.17 & Person & -0.15\\
         Covered & -0.07 & Playing& -0.13 & Girl & -0.17 & Table & -0.13 & Baseball& -0.12 & Wearing & -0.16 & City & -0.15\\
         \hdashline
         Pizza & 0.12 & Sign & 0.17 & Walking & 0.16 & Beach & 0.15 & Desk & 0.12 & Umbrella & 0.15 & Standing & 0.14\\
         Umbrella & 0.13 & Desk & 0.18 & Clock & 0.17 & Group & 0.20 & Cat & 0.13 & Player & 0.16 & Their & 0.15\\ 
         Tennis & 0.14 & Boy & 0.18 & Train & 0.17 & Yellow & 0.20 & Food & 0.14 & Sink & 0.17 & Computer & 0.16\\ 
         Clock & 0.18 & Umbrella & 0.20 & Sink & 0.19 & Umbrella & 0.21 & Men & 0.14 & Road & 0.22 & Sign & 0.18\\ 
         Cat & 0.20 & People & 0.20 & Food & 0.19 & Sitting & 0.25 & Sign & 0.21 & Green & 0.23 & Player & 0.19 
    \end{tabular}
    \label{tbl:coefs}
\end{table*}
\newpage
\section{Vocabulary differences coefficients}
\label{sec:appendix_vocab}
In Sec.~\ref{sec:experiments_normalized} we explore the different word choices in the captions describing \texttt{lighter} and \texttt{darker} images. We provide the most predictive words for \texttt{lighter} and \texttt{darker} across the manual captions and the automatically generated captions in Tbl.~\ref{tbl:coefs}.

\end{document}